\documentclass{sig-alternate}

\usepackage{multicol}
\usepackage{url}

\begin{document}
%

\title{Expresso: A User-Friendly GUI for Designing, Training and Exploring
Convolutional Neural Networks}
%
%
%
%
%

\numberofauthors{2} 
%
%
%
\author{
\alignauthor Ravi Kiran Sarvadevabhatla\\
       \email{ravikiran@ssl.serc.iisc.in}
\alignauthor R. Venkatesh Babu\\
       \email{venky@serc.iisc.in} 			
\end{tabular}\newline\begin{tabular}{c}
       \affaddr{Video Analytics Lab, Supercomputer Education and Research Centre}\\
       \affaddr{Indian Institute of Science}\\
       \affaddr{Bangalore, INDIA - 560012}			
}


\maketitle

\begin{abstract}
 With a view to provide a user-friendly interface for designing, training and developing deep learning frameworks,  we have developed Expresso, a GUI tool written in Python. Expresso is built atop Caffe, the open-source, prize-winning framework popularly used to develop Convolutional Neural Networks. Expresso provides a convenient wizard-like graphical interface which guides the user through various common scenarios -- data import, construction and training of deep networks, performing various experiments, analyzing and visualizing the results of these experiments. The multi-threaded nature of Expresso enables concurrent execution and notification of events related to the aforementioned scenarios. The GUI sub-components and inter-component interfaces in Expresso have been designed with extensibility in mind. We believe Expresso's flexibility and ease of use will come in handy to researchers, newcomers and seasoned alike, in their explorations related to deep learning. 
\end{abstract}




\vspace{4cm}
\section{Introduction}

In the recent years, deep learning has established itself as a disruptive framework, raising the state-of-the-art for challenging problems related to signal recognition and analysis~\cite{amund}. In particular, Convolutional Neural Networks (CNNs), a category of deep learning networks, have shown impressive performance on challenging image recognition datasets~\cite{DonahueJVHZTD13, girshick2014rcnn, krizhevsky2012imagenet}. The rapid adoption of CNNs has been made possible to a large extent by the availability of ready-to-use software for designing CNNs and related data processing. Of these softwares, the Caffe framework~\cite{jia2014caffe} is one of the most popular. The detailed nature of source and usage documentation along with an active user forum has contributed immensely to Caffe's popularity. The fact that Caffe was the prize-winning entry in the competitive ACMMM--2014 Open Source Software Competition further promotes its credibility.

\begin{figure}[t!]
  \centering
  \includegraphics[height=5cm,width=0.48\textwidth]{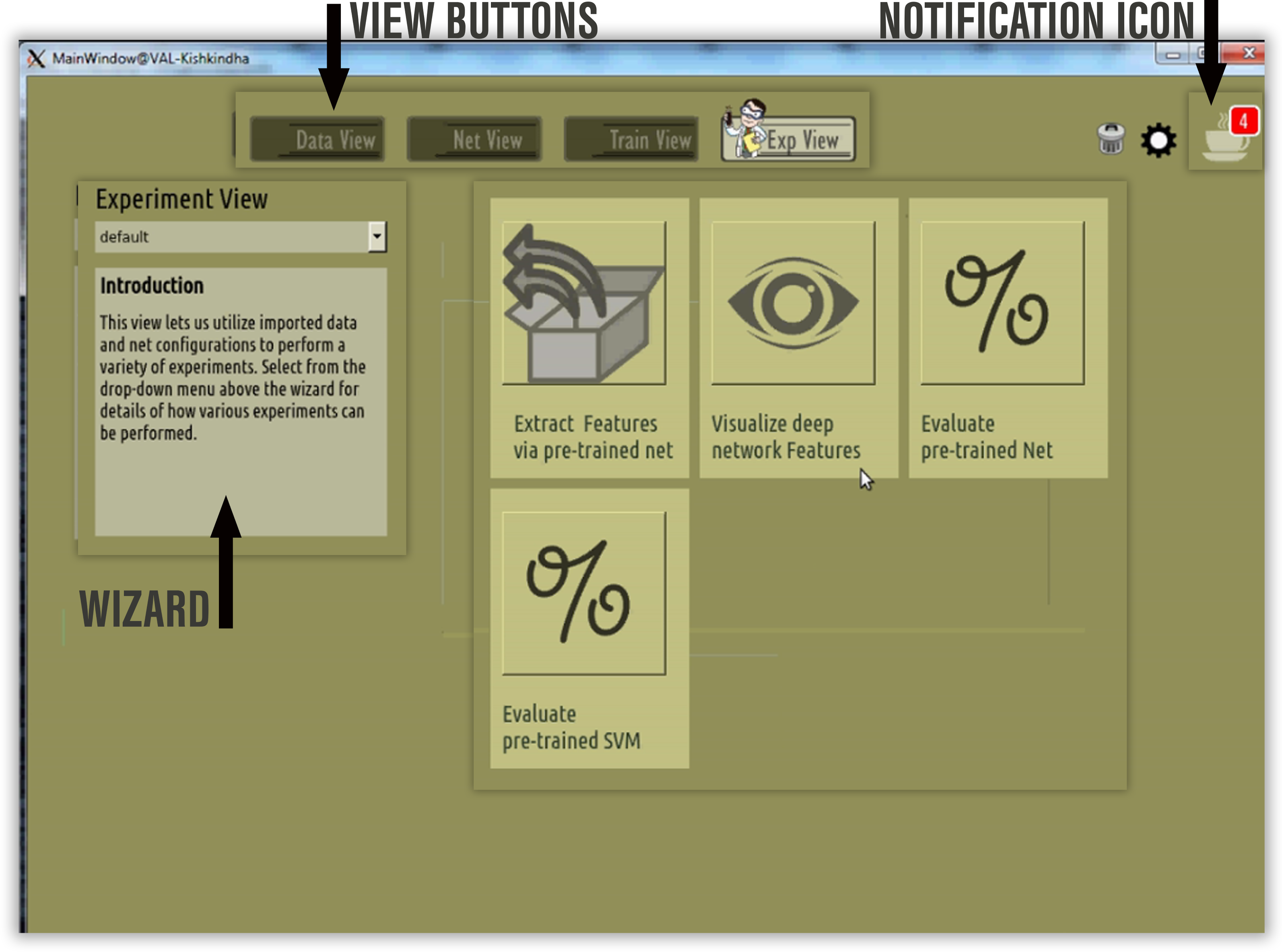}
  \caption{Expresso's main screen}
	\label{fig:expresso}
\end{figure}

The Caffe framework is primarily command-line based. To supplement Caffe's functionality and expand its ease of use, particularly for newcomers, we have developed Expresso. Expresso is a GUI tool written in Python which uses Caffe as the back-end framework for CNNs. The project page (\url{http://val.serc.iisc.ernet.in/expresso}) provides an overview of Expresso's main features and  information regarding installation and usage.

\vspace{5cm}
\section{Salient Features of Expresso}
\begin{itemize}
  \item Expresso provides a convenient wizard-like interface to contextually guide the user during common scenarios such as data import, design and training of CNNs, performing experiments and visualizing their results (Figure \ref{fig:expresso}). The wizard interface avoids the cognitive load involved in switching between the GUI and reading documentation from another interface (e.g. PDF file or browser window).
	\item Expresso's multi-threaded architecture enables asynchronous, concurrent execution of tasks in the aforementioned scenarios. 
	\item Important events that occur related to various tasks (task initialization, progress and termination) are conveyed to users via a notification system similar to that present in social media user-interfaces (Figure \ref{fig:expresso}). 
	\item The modular nature of GUI sub-components and inter-component interfaces in Expresso have been designed with ease of extensibility in mind.		
\end{itemize}

\section{The Expresso framework}
\label{sec:framework}
Expresso operates primarily as four views - \textsc{Data, Net, Train, Experiment}. These ``views" reflect the predominant use-case scenarios involving CNNs. In Expresso's GUI, each view and associated interface elements are rendered in a uniform color livery for ease of visual grouping (Figures \ref{fig:expresso},\ref{fig:netview},\ref{fig:trainview}). Next, we describe these views.

\begin{figure}[t!]
  \centering
  \includegraphics[height=5cm,width=0.48\textwidth]{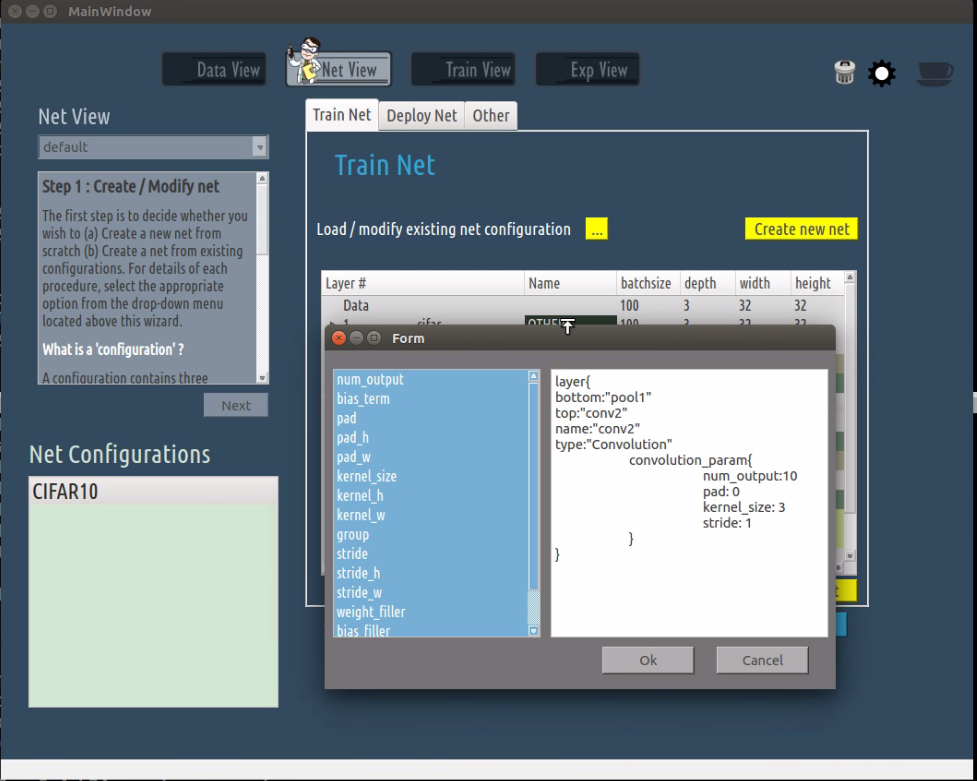}
  \caption{Using the intelli-sense editor for creating and modifying layers of a deep net}
	\label{fig:editor}
\end{figure}

\subsection{\textsc{Data} view}
\label{sec:data}
This view is meant for import of input data and typically forms the first step for most tasks. The import interface is quite flexible and enables data import in a variety of input formats\footnote{Currently, \texttt{Text,LevelDB,.mat,HDF5,Folder} formats are supported.} -- starting from the basic ``all-images-in-the-folder" up to a sophisticated, high-capacity format such as HDF5. Auxiliary information typically associated with data such as class labels can also be concurrently imported. 
A summary of salient details related to the imported data are presented in a list-based view upon import. A rudimentary image browser is also provided which can be used for a quick preview of the imported data. In addition, the imported data can also be exported to other formats. Currently, \texttt{.mat} and \texttt{HDF5} are supported. A common requirement involves split of data into training and testing sets. This functionality is also provided in the \textsc{Data} view.

Another interesting feature is that the back-end framework related to \textsc{Data} view also features a parser. This parser can be utilized to quickly extend the import interface to load data in formats not provided yet in \textsc{Data} view.

\subsection{\textsc{Net} view}
\label{sec:net}
The \textsc{Net} view is utilized for constructing deep neural network architectures (nets) -- either from scratch or by modification of existing nets. The net construction phase requires specification of a training `prototxt' file whose contents are interpreted by Caffe as the specification for network architecture. A related text file, usually referred to as the `deploy' prototxt file is associated with the usage of the network for testing and other experimentation. Optionally, a solver file is also present which contains various auxiliary parameter settings. The purpose of the \textsc{Net} view is to generate all the files required to specify a network in Caffe-compatible format.
\begin{figure}[t!]
  \centering
  \includegraphics[height=5cm,width=0.48\textwidth]{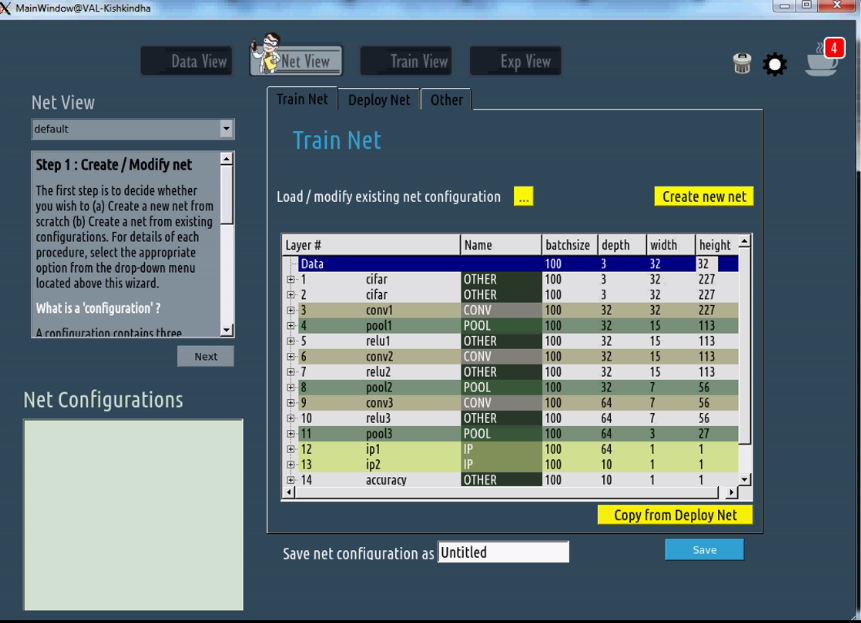}
  \caption{Viewing the layers of a pre-trained deep net. The net's layers are color-coded by type.}
	\label{fig:netview}
\end{figure}

In \textsc{Net} view, two net construction options exist.

\subsubsection{Constructing the net from scratch}

Once the option is chosen to construct the net from scratch, the layers which are associated with data are added automatically. Network layers of various common types can be subsequently added below the data layers. The back-end framework automatically calculates the sizes of responses for the resulting layers. 

The settings for each layer can be edited using a contextual smart-edit interface. The various properties of a layer are typically specified in programming-language style `scopes'. The smart-edit interface automatically provides layer description options depending on the position of the edit cursor within the layer description scope (see Figure \ref{fig:editor}). As each layer is completed, the net design so far can be viewed. The graphical nature of both this process and the smart-edit interface provides an easy-to-use approach for CNN net design.

At the end of the process, the resulting net, typically called the `training net' forms the basis for the related `deploy net' prototxt file and therefore, the latter is initialized as a slightly modified copy of the `training net' prototxt file. All the aforementioned files, related to the newly constructed net can be saved into a net configuration. This freshly created net configuration can be trained by following the steps detailed in \textsc{Training} view (Section \ref{sec:training}).

\subsubsection{Create modified version of an existing net}

Alternately, if the objective is to create a modified version of an existing net, a net needs to be chosen from the list of available net configurations. Making this choice pre-loads the associated prototxt files (train, deploy and optionally, the solver). Having done so, the subsequent modifications to the net can be done as described earlier. 

A helpful feature of the \textsc{Net} view is that the layers of a net are displayed in color-coded fashion for ease of visualization (see Figure \ref{fig:netview}). Another important feature is that the created nets are Caffe-compatible files. Therefore, the files associated with a net configuration (train prototxt file, deploy prototxt file etc.) can be utilized directly with Caffe, outside of Expresso as well.

\subsection{\textsc{Training} view}
\label{sec:training}
The \textsc{Training} view is utilized to train the nets with user-specified data. The work flow proceeds as follows :
 
\begin{enumerate}
	\item Select a net for training : This choice is made from a list of previously created net configurations whose location information is cached. Alternately, unlisted nets can be added for faster look-up in future sessions. 
	\item Set training parameters : If the net configuration does not have a solver, one is defined at this point. The solver is typically associated with a number of associated parameters. These parameters are set next, with an option for modifying and saving them if the solver already exists (see Figure \ref{fig:trainview}). 
	\item Data association : The data to be used for training is chosen next. 
	\item Choosing net vs external training : Two kinds of training are commonplace with CNNs. The first kind involves training the net from scratch. The second kind involves extracting features obtained by passing training data through a pre-trained net and tapping the output at pre-specified layer(s) of the net -- a task which can be done from the \textsc{Experiment} view. The extracted features are used to train an external classifier such as a Support Vector Machine (SVM)~\cite{libsvm,liblinear}. If such a choice is made, the associated parameters (type of SVM kernel, related parameter settings) are specified.
\item Commence training : Pressing the `Finish' button in this step kick starts the net training process. A notification panel provides information related to progress of training such as number of epochs completed, estimated time to completion and training error plots. 
\end{enumerate}

At the end of training, a model file is created regardless of training choice (net-based or external-classifier-based).

An important feature in the \textsc{Train} view is the ability to stop a particular training task. This is particularly useful if the training is not proceeding satisfactorily (e.g. too slowly or has an unacceptable loss function magnitude).

\begin{figure}[t!]
  \centering
  \includegraphics[height=5cm,width=0.48\textwidth]{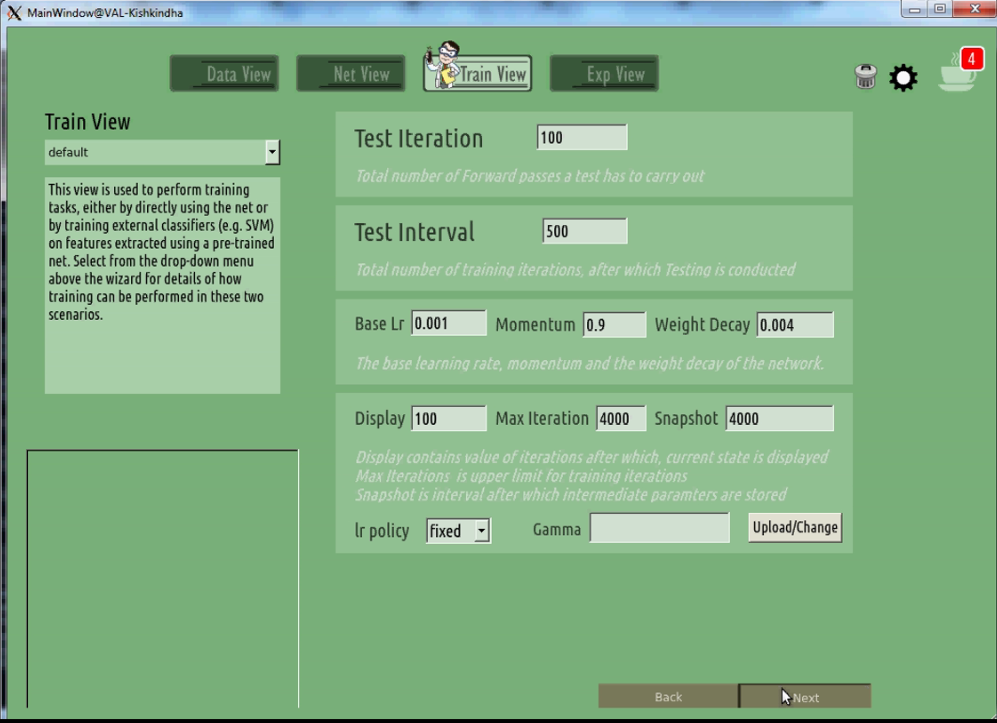}
  \caption{Interface to upload and adjust training parameters in Train view.}
	\label{fig:trainview}
\end{figure}

\subsection{\textsc{Experiment} view}
\label{sec:experiment}
The \textsc{Experiment} view can be used for three important tasks, each of which is described below.

\subsubsection{Feature extraction using pre-trained nets}

When data is provided to a pre-trained CNN, the corresponding output from CNN layers has been shown to act as informative and discriminative feature representation~\cite{girshick2014rcnn} of the data. To facilitate feature extraction, the user is asked to first choose a pre-trained net and then choose the layer(s) of interest. Having made the choice, the user then selects the data to be provided to the net. The subsequent step of feature extraction results in feature data which can be stored in a variety of data formats, such as the ones compatible with the popular libsvm software~\cite{libsvm}.

\subsubsection{Visualizing feature data}

Features from certain layers of CNNs tend to preserve and highlight various spatial aspects of input data~\cite{DonahueJVHZTD13}. To help visualize such features, an option is provided wherein features extracted previously can be selected. Such visualization can lead to a better understanding of both the data and evolution of features at various layers of the net (see Figure \ref{fig:featviz}).

\subsubsection{Testing pre-trained models}

To determine the performance of the CNN, previously unseen data is processed by a trained CNN and the results are compared against ground truth. For this process, the net configuration is chosen first. Subsequently, the test data is selected. Once the testing concludes, various standard performance metrics such as accuracy (global and per-class) and the confusion matrix are displayed. As with \textsc{Training} view, testing using external classifiers (SVM) can also be performed on features extracted from CNN layers. 

\begin{figure}[t!]
  \centering
  \includegraphics[height=5cm,width=0.48\textwidth]{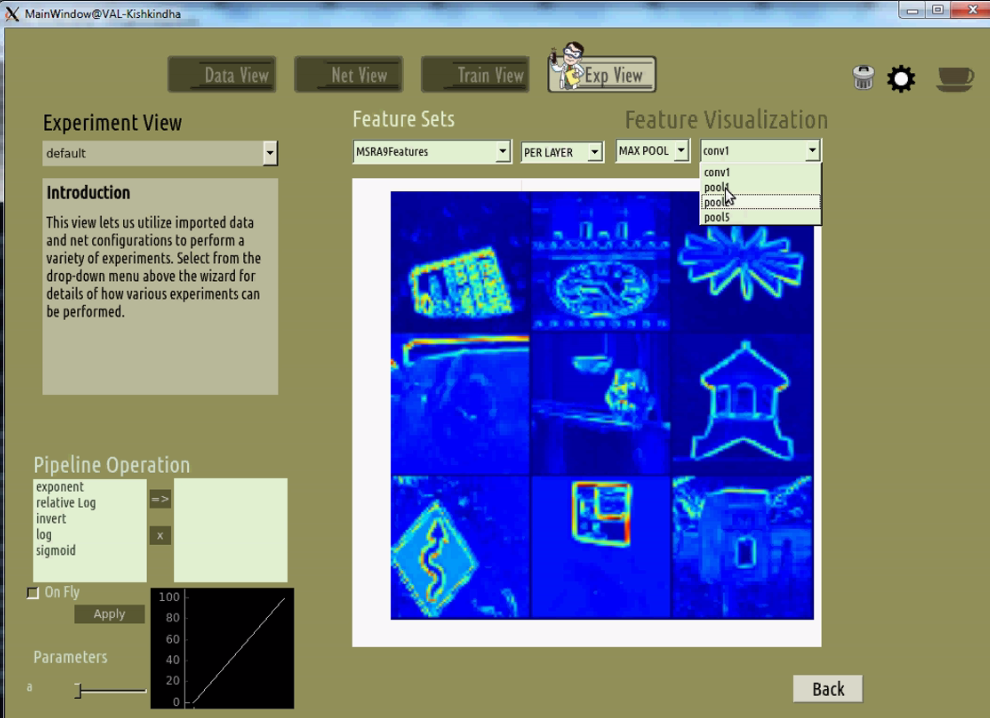}
  \caption{Visualizing layer features of a deep network}
	\label{fig:featviz}
\end{figure}

\section{ADDITIONAL NOTES}
Expresso is written in Python and has been developed on machines running Linux Ubuntu 14.04. The BSD-licensed source code for Expresso can be accessed on Github at \url{https://github.com/val-iisc/expresso}. Expresso's project page (\url{http://val.serc.iisc.ernet.in/expresso}) provides instructions and information regarding installation and usage. An illustrated set of tutorials have also been provided with Expresso in text and narrated video formats. These tutorials, covering typical use scenarios, assist the user in gaining familiarity with Expresso's functionality. 

We also note that since Expresso has been built atop Caffe, future improvements and extensions to the latter are readily conferred to Expresso. 

\section{CONCLUSION}
In this paper, we have presented a brief overview of the first version of Expresso -- a GUI tool for designing, training and using Convolutional Neural Networks. In the spirit of open-source movement, we hope that users will not only use the tool, but also extend it, thereby making Expresso the go-to GUI based tool for exploring deep learning.

\section{ACKNOWLEDGEMENT}
We gratefully acknowledge the support of NVIDIA Corporation for their donation of K40 GPU as a hardware grant. We would like to thank the members of Video Analytics Lab, IISc (\url{http://val.serc.iisc.ernet.in}) for their feedback during the development of Expresso. Finally, we would like to acknowledge the creators of Caffe for providing the shoulders of giants on which we stand.

\bibliographystyle{abbrv}
\bibliography{sigproc}  
\balancecolumns 
\end{document}